\documentclass[letterpaper, 10 pt, conference]{ieeeconf}  %

\IEEEoverridecommandlockouts
\usepackage[T1]{fontenc}
\usepackage[utf8]{inputenc}
\usepackage{csquotes}
\usepackage[english]{babel}
\usepackage[export]{adjustbox}
\usepackage[acronym]{glossaries}
\usepackage{flushend}

\let\labelindent\relax %
\usepackage{enumitem}

\usepackage[algoruled,linesnumbered]{algorithm2e}
\DontPrintSemicolon

\usepackage{hyperref}
\hypersetup{
    colorlinks=true,
    linkcolor=black,
    citecolor=black,
    filecolor=black,
    urlcolor=black,
}
\usepackage[capitalise]{cleveref}
\usepackage{orcidlink}

\usepackage{graphicx}
\graphicspath{{figures/}}

\usepackage[range-units=single]{siunitx}
\usepackage[style=ieee,
            doi=true,
            url=true,
            mincitenames=1,
            maxcitenames=1,
            minbibnames=6,
            maxbibnames=6,
            backend=biber]{biblatex}  %
\usepackage{customAcronyms}
\usepackage{customCommands}
\usepackage{customMath}
\usepackage{customStyles}

\usepackage[shortcuts]{extdash}

\usepackage{textcomp}
\usepackage[absolute]{textpos}
\usepackage{setspace}

\TPMargin{5pt}

\newcommand*{\presentationfont}{\fontfamily{lmss}\selectfont}
\newcommand{\presentationinformation}[2]{
\TPshowboxesfalse 
\begin{textblock}{0.82}(0.02,0.01)
    \vspace{2mm}
    \noindent{\bfseries{\presentationfont{\footnotesize{#1 \\
                                                        DOI: \href{http://dx.doi.org/#2}{#2}}}}}
\end{textblock}
}

\newcommand{\copyrightstatement}[1]{
    \begin{textblock}{0.825}(0.089,0.94)
        \setstretch{0.65}%
        \noindent
        \begin{minipage}{\linewidth}
            {\fontsize{6.5pt}{8pt}\selectfont
                \parbox{\dimexpr\linewidth-0mm}{%
                    \copyright{} #1 IEEE.
                    Personal use of this material is permitted.
                    Permission from IEEE must be obtained for all other uses, in any current or future media,
                    including reprinting/republishing this material for advertising or promotional purposes,
                    creating new collective works, for resale or redistribution to servers or lists,
                    or reuse of any copyrighted component of this work in other works.
                }
            }
            \vspace{-1mm} %
        \end{minipage}
    \end{textblock}
}

\title{%
    Better Safe Than Sorry: Enhancing Arbitration Graphs for\\ Safe and Robust Autonomous Decision-Making
}

\author{
    Piotr Spieker \orcidlink{0000-0002-0449-3741} $^{1*}$,
    Nick Le Large \orcidlink{0009-0006-5191-9043} $^{2*}$ and
    Martin Lauer \orcidlink{0000-0003-4414-5722} $^{2}$%
    \thanks{
        $^{1}$Piotr Spieker, né Orzechowski, is with dotscene GmbH, Freiburg, Germany
    }
    \thanks{
        $^{2}$Nick Le Large and Martin Lauer are with the Institute of Measurement and Control Systems, Karlsruhe Institute of Technology (KIT),
        Karlsruhe, Germany
    }
    \thanks{
        $^{*}$Authors contributed equally to this work
    }
}

\begin{document}

\presentationinformation{Presented at 2025 IEEE International Conference on Systems, Man, and Cybernetics (SMC)}{10.1109/SMC58881.2025.11343055}
\maketitle
\copyrightstatement{2025}

\thispagestyle{empty}
\pagestyle{empty}

\begin{abstract}

This paper introduces an extension to the \linebreak
arbitration graph framework designed to enhance the safety and robustness of autonomous systems in complex, dynamic environments.
Building on the flexibility and scalability of arbitration graphs, the proposed method incorporates a verification step and structured fallback layers in the decision-making process.
This ensures that only verified and safe commands are executed while enabling graceful degradation in the presence of unexpected faults or bugs.
The approach is demonstrated using a Pac-Man simulation and further validated in the context of autonomous driving,
where it shows significant reductions in accident risk and improvements in overall system safety.
The bottom-up design of arbitration graphs allows for an incremental integration of new behavior components.
The extension presented in this work enables the integration of experimental or immature behavior components
while maintaining system safety by clearly and precisely defining the conditions under which behaviors are considered safe.
The proposed method is implemented as a ready to use header-only C++ library, published under the MIT License.
Together with the Pac-Man demo, it is available at \linebreak
\href{https://github.com/KIT-MRT/arbitration_graphs}{\nolinkurl{github.com/KIT-MRT/arbitration_graphs}}.

\end{abstract}
\section{Introduction}

\subsection{Motivation}

Behavior planning and decision-making are crucial for robots to operate autonomously in dynamic environments, ensuring to achieve their goals while adapting to changes and uncertainties.
Key to reliable operation in fields like mobile, industrial, or service robotics is ensuring safety and robustness in these processes.

Arbitration graphs, hierarchical behavior models, manage complex decision-making
by allowing integration of diverse methods while ensuring scalability, maintainability, and transparency.
However, real-world complexities challenge the safety and robustness of such systems.

This paper aims to enhance arbitration graph safety and robustness by identifying and handling erroneous or unsafe behavior commands at runtime.

\begin{figure}
    \centering
    \includegraphics[width=0.7\columnwidth]{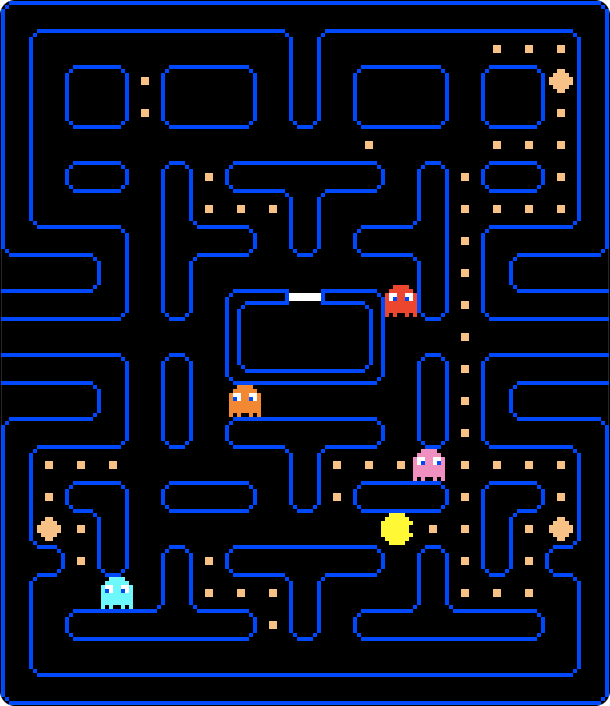}
    \caption{The Pac-Man simulation used to demonstrate the presented extension to the arbitration graph framework.
    Leveraging the framework's flexibility and scalability,
    we incorporate a verification step and fallback layers
    to ensure robust and safe decision-making.}
    \label{fig:entt-pacman}
\end{figure}

\subsection{State of the art}
Behavioral decision-making includes both monothematic methods and generic architectures.

End-to-end machine learning approaches learn the entire process from sensor data to commands, requiring extensive data and computational power.
Due to their highly integrated nature, it is challenging to interpret or influence the resulting behavior directly.

Traditional architectures like \glspl{FSM} allow situational planning but scale poorly with complexity. Behavior-based methods, derived from Brooks' subsumption architecture, evolved into \glspl{BT}, among others. Popularized by their use in gaming, they are also applied in robotics. These provide a hierarchical decision making structure, offering modularity and responsiveness but becoming cumbersome with extensive conditions.

Arbitration graphs, combining subsumption and object-oriented programming, enhance reusability and system clarity through modularity and functional decomposition. Used in robotic soccer \cite{lauerCognitiveConceptsAutonomous2010} and automated driving \cite{orzechowskiDecisionMakingAutomatedVehicles2020}, these graphs employ behavior components to interpret situations and plan actions, while arbitrators select the most suitable behaviors.

\subsection{Contributions}

\begin{description}[align=left]
    \item[Verification Logic] We extend the arbitration process to ensure that only verified behaviors are executed.
    \item[Fallback Logic] We introduce fallback options for cases where behavior commands fail verification.
    \item[Application] We validate the safety concept in autonomous driving simulations, demonstrating reduced accident risk and improved safety.
\end{description}

\section{Fundamentals}

\subsection{Decision-Making}

Decision-making is crucial in the planning module of a robot, determining commands based on the current situation.
Methods range from graph-based techniques like A*, PRM*, and RRT* to probabilistic and machine learning methods.
This work, however, focuses on rule-based methods, such as \glspl{FSM}, \glspl{BT}, or arbitration graphs, which address decision-making through discrete state or mode transitions.
Refer to \cite{schwartingPlanningDecisionMakingAutonomous2018,yurtseverSurveyAutonomousDriving2020,gammellAsymptoticallyOptimalSamplingBased2021}
for a detailed overview of available approaches.

\textbf{\glspl{FSM}}, first used in hardware design and theoretical computer science,
represent behavior modes with transitions triggered by events \cite{wagnerModelingSoftwareFinite2006}.
Despite their simplicity, \glspl{FSM} scale poorly and are difficult to modify due to the amount of transitions increasing exponentially with the number of states.

\textbf{\glspl{BT}}, initially designed for game development \cite{iovinoProgrammingEffortRequired2022},
have been increasingly used in robotics since 2012 \cite{bagnellIntegratedSystemAutonomous2012,ogrenIncreasingModularityUAV2012}.
They separate behavior decision-making from execution using a tree structure \cite{colledanchiseBehaviorTreesRobotics2018}.
Internal nodes determine selection mechanisms, while leaves describe behaviors and conditions.
Evaluated at a fixed frequency, nodes return their status as running, completed, or failed.
Control flow nodes decide on further evaluations.
Condition nodes check if their conditions are met without affecting the environment, while action nodes execute behaviors and return their status.

\glspl{BT} generalize many architectures, such as hierarchical \glspl{FSM} and decision trees \cite{colledanchiseHowBehaviorTrees2017} excelling in modularity, hierarchical organization, reusability, responsiveness, and interpretability \cite{colledanchiseBehaviorTreesRobotics2018}.
Their flexibility allows reuse of individual behaviors \cite{bagnellIntegratedSystemAutonomous2012}.
The selection mechanism is intuitive and easy to follow during operation.
However, extensive preconditions can make representations cumbersome, and safety as well as reliability depend significantly on node arrangement.
These drawbacks are addressed by arbitration graphs.

\textbf{Arbitration graphs} originated in the context of robot soccer \cite{lauerCognitiveConceptsAutonomous2010},
integrating ideas from Brooks' behavior-based subsumption \cite{brooksRobustLayeredControl1986},
knowledge-based architectures like Belief-Desire-Intention (BDI) \cite{raoAbstractArchitectureRational1992},
and programming paradigms such as object-oriented programming \cite{stefikObjectOrientedProgrammingThemes1985}.

This modular framework is characterized by clear interfaces for transparent decision-making,
using atomic behavior components to represent simple abilities and behaviors.
These modules are combined using arbitrators to create complex system behaviors.

The input to a behavior component is the current situation~$\situation$, provided as sensor data or an interpreted environment model.
If its preconditions are met, the $\invCond$ condition indicates that the behavior component is applicable in the current situation~$\situation$.
In this case, the higher-level instance (i.e., the arbitrator) can instruct the behavior component to compute a command~$\command$.
The currently active behavior component additionally uses the $\comCond$ condition to indicate that its behavior can be continued.
Consequently, the calling instance does not need to know the prerequisites for executing the command returned by the behavior.

Generic arbitrators combine behavior components $\options~=~\left< o_0, o_1, \dots \right>$,
filter out the applicable subset $\applicableOptions \subseteq \options$ using their invocation and commitment conditions,
and select the best applicable option~$a^*$ for execution.
Arbitration schemes include priority-based, sequence-based, cost-based, and random.
Due to inheritance and a shared interface, arbitrators can include both behavior components and other arbitrators, enabling a hierarchical architecture.

\textbf{Comparing Arbitration graphs to BTs}, both appear similar at first glance, but differ in several key aspects:

\gls{BT} nodes return execution status, while behavior components in arbitration graphs return commands.
While the former adds more flexibility to each node's actuator interfaces,
the latter focuses on a control theory motivated interface ${f(\situation) \to \command}$ %
allowing the command to be verified by each arbitrator before executing it in a down-stream module.

In \glspl{BT}, preconditions are implemented by condition nodes distributed throughout the tree.
In contrast, arbitration graphs require behavior components to define their own preconditions.
This tight coupling makes robustness and safety less dependent on the node arrangement.
\glspl{BT} rely on each node to decide on its success or failure,
which can lead to safety and reliability issues if not carefully managed.
Arbitration graphs, as enhanced in this paper, integrate safety into the selection mechanism,
using node-independent verifiers.
This functional decomposition reduces the burden on behavior engineers and
allows an easy integration of unsafe, e.g.\ learning-based, behavior components.

\subsection{Fault-tolerant/Robust Systems}

In automated systems, both hardware and software issues can compromise performance and safety.
Causes include programming errors and runtime issues such as optimization problems, making error diagnosis and treatment crucial in system design and during runtime.

Reliable systems aim to maintain performance
despite potential errors, using metrics like error probability, mean lifespan, failure rate, and availability \cite{echtleFehlertoleranzverfahren1990}.
Terminology varies, but disturbances (\enquote{faults}) can lead to errors, potentially causing system failures.

Reliability measures include error prevention, removal, tolerance, and prediction \cite{dubrovaFaultTolerantDesign2013}.
Prevention and removal focus on design and development, while tolerance involves detecting and preventing operational errors.
Prediction estimates future failures.
Error tolerance involves diagnosing and handling errors—restoring faulty components,
computing correct results despite faults, or removing faulty components \cite{echtleFehlertoleranzverfahren1990}.

\subsection{Pac-Man}

\begin{figure}
    \centering
    \includegraphics[width=\columnwidth]{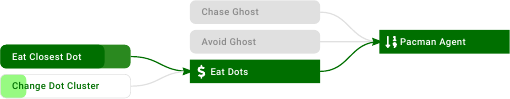}
    \caption{A basic arbitration graph for playing Pac-Man.}
    \label{fig:pacman-arbitrator-base}
\end{figure}

Pac-Man, the 1980 arcade game, involves navigating a maze to eat dots while avoiding ghosts.
Depicted in \cref{fig:entt-pacman} is an open-source implementation\footnote{\url{github.com/indianakernick/EnTT-Pacman}}
of the game that is being used to demonstrate the proposed method.
We split the behavior of Pac-Man into the following behavioral components:
\begin{description}[align=left]
    \item[Avoid Ghosts] Pac-Man tries to increase the distance to the ghosts to avoid being eaten.
    \item[Chase Ghosts] After consuming an energizer, Pac-Man might try to eat the ghosts for extra points.
    \item[Eat Closest Dot] Pac-Man moves towards the dot that is closest to him.
    \item[Change Dot Cluster] Pac-Man moves towards another area with a higher dot density.
\end{description}

The corresponding arbitration graph is shown in \cref{fig:pacman-arbitrator-base}.

\section{Safe Arbitration Graphs}

Designing autonomous systems ideally involves ensuring that each behavior component operates reliably under all potential conditions.
However, achieving this level of reliability is often impractical, especially when deploying these systems in complex and dynamic environments.
Real-world applications might involve numerous behaviors, implemented by different teams and using various methods.
With increasing complexity, the likelihood of bugs and inconsistencies grows, making it challenging to ensure the overall safety and robustness of such systems.

The responsibility for safety in autonomous platforms is a critical consideration.
Typically, each developer is responsible for their behavior component.
Additionally, when using hierarchical approaches such as \glspl{BT},
    a system engineer is in charge of integrating these behavior components into a coherent system,
    bearing the overall responsibility for operational safety.
This responsibility is manageable when components are simple and isolated.
However, it does not scale well with increasingly complex architectures.

To address these challenges, we propose a safety concept embedded directly into the arbitration graph framework.
With this approach, we aim to reduce the burden on both behavior component developers and system engineers
by shifting the responsibility for the platform's safety to the verification step within the decision-making framework.
This ensures that even in the presence of unreliable or unsafe behavior components, the system as a whole remains robust and safe.

In this section, we will explore methods for detecting and mitigating unsafe and unreliable behavior components, with the following safety goals in mind:

\begin{itemize}
    \item The system ensures that a safety action is taken when operating conditions exceed its designed capabilities.
    \item The system is robust against failures of behavior components.
    \item The system prevents invalid behavior commands.
    \item The system avoids risky behavior commands.
\end{itemize}

\subsection{\textbf{Detect} Unsafe \& Unreliable Behaviors}
\begin{algorithm}
  \SetKwFunction{BestOption}{BestOption}%
  \SetKwProg{Fn}{function}{}{end}

  \Fn{\BestOption{situation $\situation$}}{\label{algo:verifying_arbitrator:bestOption}
    filter applicable options $\applicableOptions \subseteq \options$\;
    sort applicable options
      $\sortedApplicableOptions = \left< a_0, a_1, \dots \right> = \text{strategy}(\applicableOptions)$\;

    \For{$a \in \sortedApplicableOptions$}{
      get command $\command_a = \getCommand_a(\situation)$\;

      verify $\verification_a = \verifier (\command_a)$\;
      \If{verification passed $\verification_a = 0$}{%
        \KwRet{$(\command_a, \verification_a)$}\;
      }
    }

    \KwRet{$(\emptyset, \text{NO\_SAFE\_OPTION})$}\;
    }
  \;
  \While{true}{
    determine current situation $\situation$\;\label{algo:verifying_arbitrator:situation}
    determine $\verifiedCommand = (\command, \verification) = $ \BestOption{$\situation$}\;
    \If{verification passed $\verification_a = 0$}{
      execute $\command$\;
    }
  }
  \caption{Generic arbitration with verification \label{algo:verifying_arbitrator}}
\end{algorithm}

\Cref{algo:verifying_arbitrator} shows the generic arbitration algorithm from \cite{lauerCognitiveConceptsAutonomous2010}
that we have extended with a verification logic.
The verifier~$\verifier$ is domain-specific and may run various checks on the behavior command.
In Pac-Man for example, the verifier could check whether the command would lead to a collision with walls or ghosts.
In autonomous driving, one could verify if the behavior output is free of collisions and respects traffic rules.
Similarly, a robotic manipulator could verify if the command is within the workspace of the robot and does not exceed joint limits.
More generally, the verifier could check if the format of the command is correct and respects the specification of the system.
Integrating the verification step into the arbitration algorithm allows for the use of different verifiers at various levels within the arbitration graph.
This enables a more fine-grained and adaptable safety concept.
For example, verifiers can be customized for specific scenarios, or computationally intensive verifiers can be reserved for higher levels of the arbitration graph.

The verification step is embedded as follows:
First, in \cref{algo:verifying_arbitrator:situation}, the current situation~$\situation$ is determined.
Based on this, the root arbitrator determines its best applicable and safe action using the \BestOption{$\situation$} function.
For this, it determines the set of options~$\applicableOptions \subseteq \options$ which are applicable in the present situation~$\situation$.
Like in the original arbitration algorithm, the options of an arbitrator can themselves be arbitrators or behavior components.
The applicable options~$\applicableOptions$ are then sorted into a descending list~$\sortedApplicableOptions$ according to the underlying strategy.
Now, for each option $a \in \sortedApplicableOptions$, it is checked whether its command $\command_a = \getCommand_a(\situation)$ withstands a verification~$\verifier (\command_a)$.
If so, this option is returned as the best applicable and safe option.
If none of the options passes the verification step, the arbitrator returns the error value $\textit{NO\_SAFE\_OPTION}$.
Otherwise, the command~$\command$ returned by the root arbitrator can be executed and considered safe given the assumptions of the verifier~$\verifier$.

\subsection{\textbf{Mitigate} Unsafe \& Unreliable Behaviors}
If a behavior component fails the verification step, the arbitration graph must handle this situation to ensure the system remains safe and robust.
Even without additional measures, the verification step improves the safety of the system by preventing unsafe commands from being executed.
If an arbitrator detects a failure and has other applicable options, it simply chooses the next best option.
However, to further increase robustness and reduce performance degradation, we make use of the bottom-up approach of arbitration graphs to add fallback layers in the form of additional behavior components.

One common approach in fault-tolerant systems is to increase the system's diversity and redundancy.
In the case of arbitration graphs this can be achieved by adding further behavior components.
A redundant behavior component is merely a duplicate instantiation of an existing behavior component.
If it is non-deterministic, the redundant behavior component might find a safe command where the original behavior component failed to do so.

In contrast, a diverse behavior component addresses the same task but with a different approach.
For example, a preferred experimental or learning-based behavior component could be complemented by a fallback that employs a more conservative yet stable method to generate the command.

Another approach is to add a behavior component that repeats or continues the last command.
This helps to mitigate short-term failures such as a behavior component failing to produce a valid command for a single time step.

Finally, a last resort behavior component should be added to the arbitration graph.
This behavior should be simple and safe, ensuring that the system always has a valid command to execute.
For example, in a mobile robot, the emergency behavior could be to stop moving and wait for further instructions.
Since this is the last resort, the emergency behavior does not need to pass the verification step.

This layered approach using multiple fallback behavior components allows the performance of the system to degrade gracefully instead of having to execute an emergency command right away.

\subsection{Example: Safe Pac-Man}

\begin{figure}
    \centering
    \includegraphics[width=0.7\columnwidth]{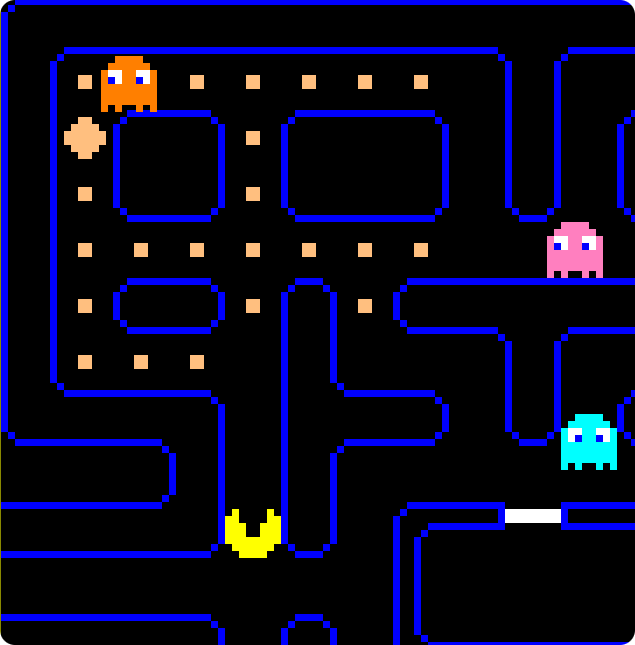}
    \caption{A scenario where \behavior{EatClosestDot} fails to produce a valid command.}
    \label{fig:pacman-scenario}
\end{figure}

\begin{figure}
    \centering
    \includegraphics[width=\columnwidth]{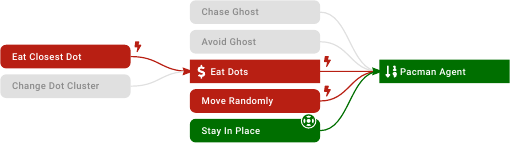}
    \caption{The extended arbitration graph with fallback layers.
        The components highlighted in red were rejected by the verifier.
        The safety buoy indicates a last resort fallback which does not need to pass verification.}
    \label{fig:pacman-arbitrator-safe}
\end{figure}

While the original arbitration graph shown in \cref{fig:pacman-arbitrator-base} might have been sufficient under normal conditions, it might fail in practice if a behavior component returns an unsafe or unreliable command.
Consider the scenario depicted in \cref{fig:pacman-scenario}.
Since the ghosts are relatively far away and there is only one dot cluster left, the only applicable behavior is \behavior{EatClosestDot}.
Ideally, this behavior has the intended effect and Pac-Man should move towards the last remaining dots.
In our hypothetical scenario, however, a bug in the underlying path planning algorithm leads to erroneous results.
Consequently, the behavior component fails to generate a valid command, causing the system to behave unpredictably or even crash.

\cref{fig:pacman-arbitrator-safe} shows the arbitration graph from \cref{fig:pacman-arbitrator-base} extended with fallback layers.
With \behavior{EatClosestDot} not returning a valid command, the system falls back to the newly added \behavior{MoveRandomly} behavior component.
Moving Pac-Man in a random direction is a simple action which might help the primary behavior components to escape a deadlock and find a new valid command.
In our scenario, however, \behavior{MoveRandomly} returns a command which would lead to a collision with a wall.
Therefore, this command is rejected by the verifier as well, and the system falls back to the last resort behavior \behavior{StayInPlace}.
While it might be impossible to complete the level with this behavior, the system remains in a predictable and safe state giving the primary behavior components a chance to recover.

Of course, in this toy example it would be possible to find the bug in \behavior{EatClosestDot} through testing.
In a real-world scenario, however, it is not feasible to test all possible situations and bugs might only manifest in specific edge cases.
The proposed safety concept explicitly handles safety within the verifiers, transferring the responsibility away from individual behavior components and their arrangement to the verifiers.
This approach allows for the integration of imperfect or experimental behavior components without compromising the overall safety of the system.

\section{Safe and Reliable Behavior Arbitration for Automated Vehicles}

Arbitration graphs can be applied to a wide range of real-world scenarios.
In \cite{orzechowskiDecisionMakingAutomatedVehicles2020},
they have been used in the context of automated driving.
There, a behavior component represents driving maneuvers of approximately $5$ to $20$ seconds
such as lane changes, parking maneuvers, or crossing intersections.

Behavior planning for automated driving comes with a variety of challenges.
The environment is highly complex and dynamic, and safety-critical decisions must be made under real-time constraints.
Among other things, the agent has to consider kinematic and dynamic constraints,
sensor limitations, traffic rules, and vehicle safety.
Therefore, verifying planned commands and providing fallback options in case of failure are crucial in this domain.

This section summarizes our work in \cite{orzechowskiVerhaltensentscheidungFuerAutomatisierte2023},
which extends \cite{orzechowskiDecisionMakingAutomatedVehicles2020} by
introducing a safety concept for arbitration graphs in the context of automated driving.

\subsection{Arbitration Graph --- \textbf{What} to do?}

\begin{figure}
    \centering
    \includegraphics[width=\columnwidth]{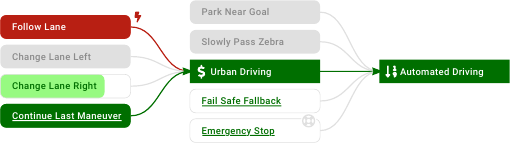}
    \caption{
        A minimalistic arbitration graph for automated driving as introduced in \cite{orzechowskiDecisionMakingAutomatedVehicles2020}, extended by fallback layers (underlined).
    }
    \label{fig:arbitration-graph-evaluation}
\end{figure}

\cref{fig:arbitration-graph-evaluation} depicts a simplified arbitration graph for automated driving
that will be used to showcase our safety concept.
It contains three behavior components for basic lateral behaviors \behavior{FollowLane}, \behavior{ChangeLaneLeft} and \behavior{ChangeLaneRight},
a parking behavior \behavior{ParkNearGoal}
and a stopping behavior \behavior{EmergencyStop}.
The arbitration graph for a real-world application would include more behavior components and arbitrators
in order to cover a wide range of driving maneuvers and scenarios.
Refer to \cite{waymo_safety_report_2020} for a detailed discussion about behavior competencies and operational design domains.

\subsection{Behavior Components --- \textbf{How} to do it?}

\begin{figure}
    \centering
    \includegraphics[width=\columnwidth]{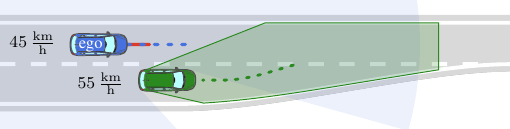}

    \caption{%
        Excerpt from the environment model with
        ego vehicle speed,
        last planned intended and fail-safe trajectory
        (blue dashed and red solid lines)
        and sensor range (blue).
        Another vehicle (green)
        with its predicted trajectory (dashed line)
        and its predicted worst-case occupancy (green).
    }
    \label{fig:environment-model}
\end{figure}

In our implementation, each behavior component has access to an environment model (situation~$\situation$) and returns a trajectory (command~$\command$).
The environment model contains information
such as the ego vehicle state, other traffic participants, and the planned route.
\cref{fig:environment-model} illustrates an example scenario with the ego vehicle in blue,
another vehicle in green, and the planned ego trajectory in dashed blue.

The $\invCond$ and $\comCond$ conditions of behavior components are derived from the operational design domain of their addressed driving maneuver.
As an example, we will examine the \behavior{ChangeLaneLeft} behavior component.
Its $\invCond$ condition is true,
if there is a lane to the left of the current lane that the ego vehicle can legally change to and the distance to objects in the adjacent lane is sufficient for a lane change.
The $\comCond$ condition is true, as long as the ego vehicle is actively changing lanes
and the target lane remains clear.
The trajectory smoothly transitions to the left lane.
See \cite{orzechowskiVerhaltensentscheidungFuerAutomatisierte2023} for more details and other behavior components for automated driving.

\begin{figure*}
  \centering
  \includegraphics[width=\textwidth]{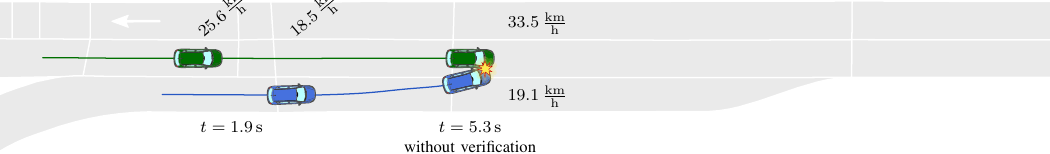}
  \caption{
      The ego vehicle intends to change lanes at $t=\SI{1.9}{\second}$,
      but another vehicle is following close and fast.
      Without verification, the too optimistic \behavior{ChangeLaneLeft} behavior is chosen,
      leading to a collision at $t=\SI{5.3}{\second}$.
      Verification using occupancy predictions prevents this
      (\cref{fig:experiments-safe-occupancies,fig:experiments-safe-arbitration-graph,fig:experiments-safe-timeline}).
  }
  \label{fig:experiments-topview}
\end{figure*}

\subsection{Verification --- What is considered \textbf{safe}?}

In order to detect unsafe and unreliable behavior commands,
various aspects need to be considered in the behavior generation process.
In our case, we verify the validity and safety of the planned trajectory.

The verifier for validity ensures that the planned trajectory fulfills the kinematic and dynamic constraints of the ego vehicle
and adheres to traffic rules.

In order to achieve vehicle safety, we
provide worst-case occupancy predictions of other traffic participants
and extend the behavior command with a fail-safe trajectory \cite{althoffSetBasedPredictionTraffic2016, tas2023decision}.
The fail-safe trajectory is designed to be safe under worst-case assumptions
and can be used in fallback layers in case of a failure or too risky situation.
As a result, each behavior component is responsible for generating both
a planned trajectory and a fail-safe trajectory (see \cref{fig:environment-model}).
The safety verifier checks if the occupancy of the fail-safe trajectory does not overlap
with the worst-case occupancies of other traffic participants,
\ie the fail-safe trajectory is collision-free under worst-case assumptions.

\begin{figure}[!t]
    \centering
    \includegraphics[width=\columnwidth]{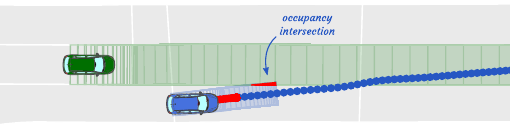}
    \caption{
        Predicted worst-case occupancies in green and
        the egos fail-safe trajectory occupancy in blue
        at $t=\SI{1.9}{\second}$.
        An overlap hints towards a risk.
    }
    \label{fig:experiments-safe-occupancies}
\end{figure}

\begin{figure}[!t]
    \centering
    \includegraphics[width=\columnwidth]{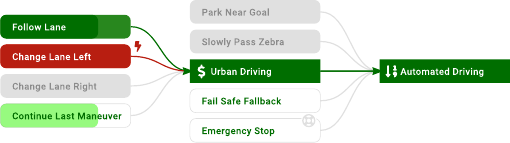}
    \caption{
        Arbitration graph with verification at $t=\SI{1.9}{\second}$.
        The safety verifier detects that the fail-safe trajectory of \behavior{ChangeLaneLeft}
        overlaps with the worst-case occupancies of the green vehicle
        (\cref{fig:experiments-safe-occupancies}).
        As a result, the \arbitrator{UrbanDriving} arbitrator choses the next best option: \behavior{FollowLane}.
    }
    \label{fig:experiments-safe-arbitration-graph}
\end{figure}

\subsection{Fallback Layers --- How to always \textbf{drive safe}?}

In order to mitigate unsafe or unreliable behaviors,
we extended the arbitration graph in \cref{fig:arbitration-graph-evaluation}
by the proposed verifiers and fallback layers
highlighted by underlined text.

In case of intermittent failures,
it is feasible to fall back to the last planned trajectory
provided by \behavior{ContinueLastManeuver},
since trajectories are planned for a given horizon.

Should the last planned trajectory no longer be suitable or safe,
\behavior{FailSafe} can be chosen to execute the fail-safe trajectory described above.

As a last resort, \behavior{EmergencyStop} can bring the vehicle to a full stop.
This may be necessary if the assumptions of the fail-safe trajectory are violated.
Since this is the last resort fallback layer,
it does not go through verification.
Hence, it is crucial that it is implemented in a simple, deterministic and reliable fashion
without the need for contextual knowledge.

\subsection{Experiments}

We validate the proposed concept in the automated driving simulator CoInCar-Sim~\cite{naumannCoInCarSimOpenSourceSimulation2018}
using handcrafted but realistic driving scenarios
from our real-world test track in Karlsruhe, Germany.
We analyze two different use cases
in \cite{orzechowskiVerhaltensentscheidungFuerAutomatisierte2023}:
\begin{description}[align=left]
    \item[Ensuring driveability]
        The verifiers and fallback layers lead to stable trajectory commands,
        even under high probability for broken trajectories.
    \item[Guaranteeing vehicle safety]
        The verifiers and fallback layers ensure collision free behavior,
        even if a behavior component provides unsafe commands.
\end{description}

Here, we focus on the latter, more critical case.
Traditional decision-making approaches select maneuver options solely based on their preconditions.
As a result, they might select unsafe maneuver options,
if the preconditions have been designed too optimistically.
\cref{fig:experiments-topview} shows such a scenario, where the ego vehicle wants to change lanes,
but another vehicle follows too closely in the target lane.
We compare the evolution of the scene with and without verification in the decision-making process.

\subsection{Results}

\begin{figure}[!t]
  \centering
  \includegraphics[width=\columnwidth]{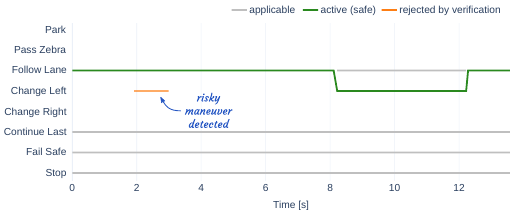}
  \caption{
      Timeline of the chosen behavior in the case with verification.
      The verifier prevents unsafe behavior commands from being executed.
      The arbitrator falls back to another safe behavior option,
      \behavior{FollowLane} in this case.
  }
  \label{fig:experiments-safe-timeline}
\end{figure} 
To provoke a risky situation,
the $\invCond$ condition of \behavior{ChangeLaneLeft} was made overly optimistic.
As a result, the arbitration graph without verification selects \behavior{ChangeLaneLeft} at $t=\SI{1.9}{\second}$
leading to a collision at $t=\SI{5.3}{\second}$ (\cref{fig:experiments-topview}).

With verification, the safety verifier detects the risk of a collision based on the worst-case occupancies of the other vehicle (\cref{fig:experiments-safe-occupancies}).
Consequently, \behavior{ChangeLaneLeft} fails the verification and
the \arbitrator{UrbanDriving} arbitrator selects \behavior{FollowLane}
as its next best option (\cref{fig:experiments-safe-arbitration-graph}).
This makes the ego vehicle slow down and not change the lane
until the green vehicle has passed (\cref{fig:experiments-safe-timeline}).
A collision has been prevented
while maintaining a smooth driving behavior.
\section{Conclusion}
This paper has presented an extension to the arbitration graph framework that focuses on improving the safety and robustness of autonomous systems operating in complex, dynamic environments.
It builds upon the strengths of arbitration graphs, which provide a flexible, scalable, and transparent decision-making framework for autonomous systems.
By embedding a verification step into the arbitrators and adding structured fallback layers to the arbitration graph,
the proposed method ensures that only verified and safe commands are executed.

The introduced method has been demonstrated using a Pac-Man simulation, where the arbitration graph successfully maintained safe operation of the agent even in the presence of unexpected faults or bugs.
Further validation has been conducted in the context of autonomous driving, where the method has demonstrated a reduction in accident risk and an improvement in system safety.
These results underline the applicability of the proposed approach to real-world problems,
confirming that arbitration graphs, when equipped with safety mechanisms,
can effectively manage the complexities and uncertainties of autonomous decision-making.

The bottom-up approach of the framework enables the incremental integration of new behavior components with diverse underlying methods into a coherent decision-making system.
The extension introduced in this work allows for the addition of new components, even if they are not fully matured or rely on experimental methods, without compromising overall system safety.
The modular structure also supports the inclusion of multiple fallback layers, ensuring graceful degradation in the face of unforeseen faults.

By explicitly defining the conditions under which a behavior component is considered safe,
the responsibility for system safety is shifted to the verifiers used by the algorithm.
This marks a crucial advancement towards safe autonomous systems, as
the overall safety of the system now primarily depends on the assumptions made by these verifiers.

\section{Acknowledgments}

The authors thank the German Federal Ministry of Education and Research (BMBF)
for being funded in the project \enquote{AUTOtech.agil} (grant 01IS22088T).

\printbibliography

\end{document}